%% file: main.tex
\documentclass[10pt,twocolumn,letterpaper]{article}

\usepackage{cvpr}              
\input{preamble}
\definecolor{cvprblue}{rgb}{0.21,0.49,0.74}
\usepackage[pagebackref,breaklinks,colorlinks,allcolors=cvprblue]{hyperref}
\usepackage{engord}

\def\confYear{}
\def\paperID{} 
\def\confName{Arxiv}
\title{DeepInv: A Novel Self-supervised Learning  Approach for Fast and Accurate  Diffusion Inversion}

\author{Ziyue Zhang\\
Xiamen University\\
{\tt\small zhang\_zi\_yue@foxmail.com}
\and
Luxi Lin\\
Xiamen University\\
\and
Xiaolin Hu\\
Xiamen University\\
\and
Chao Chang\\
National University of Defense Technology\\
\and
Huaixi Wang\\
National University of Defense Technology\\
\and
Yiyi Zhou\\
Xiamen University\\
\and
Rongrong Ji\\
Xiamen University\\
{\tt\small rrji@xmu.edu.cn}
}
\begin{document}
\maketitle
\input{sec/0_abstract}    
\input{sec/1_intro}

\input{sec/2_related}

\input{sec/3_pre}
\input{sec/4_method}

\input{sec/5_exp}

\input{sec/6_conclu}
{
    \small
    \bibliographystyle{ieeenat_fullname}
    \bibliography{main}
}


\end{document}

%% file: sec/0_abstract.tex
\begin{abstract}
Diffusion inversion is a task of recovering the noise of an image in a diffusion model, which is vital for controllable diffusion image editing. At present, diffusion inversion still remains a challenging task due to the lack of viable supervision signals. Thus, most existing methods resort to approximation-based solutions, which however are often at the cost of performance or efficiency. To remedy these shortcomings, we propose a novel self-supervised diffusion inversion approach in this paper, termed \emph{Deep Inversion} (DeepInv). Instead of requiring  ground-truth noise annotations, we introduce a self-supervised objective as well as a data augmentation strategy to generate high-quality pseudo noises from real images without manual intervention. Based on these two innovative designs, DeepInv is also equipped with an iterative and multi-scale training regime to train a parameterized inversion solver, thereby achieving the fast and accurate image-to-noise mapping. To the best of our knowledge, this is the first attempt of presenting a trainable solver to predict inversion noise step by step. The extensive experiments show that our DeepInv can achieve much better performance and inference speed than the compared methods, \emph{e.g.}, +40.435\% SSIM than EasyInv and +9887.5\% speed than ReNoise on COCO dataset. Moreover, our careful designs of trainable solvers can also provide insights to the community. Codes and model parameters will be released in https://github.com/potato-kitty/DeepInv.
\end{abstract}

\begin{figure}[!t]
  \centering
  \includegraphics[width=1\linewidth]{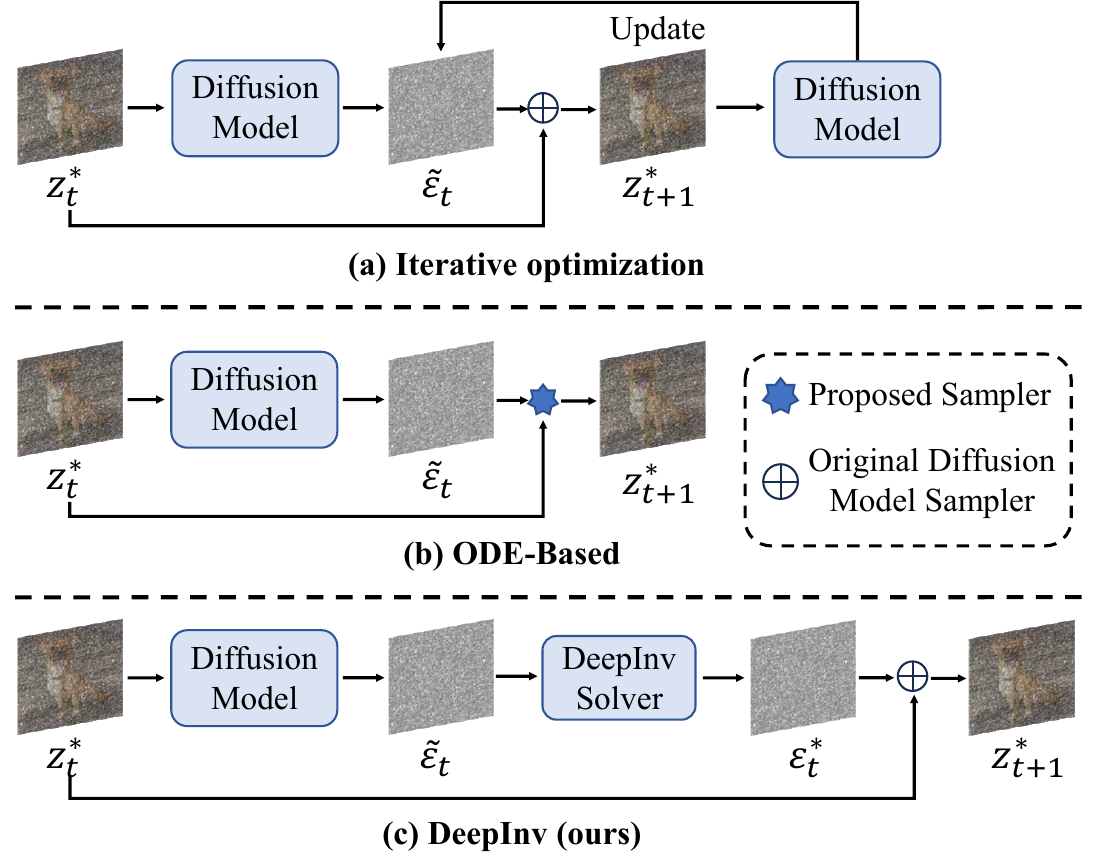}
  \caption{Illustrations of our DeepInv and previous inversion strategies. (a) Iterative optimization methods~\cite{pan2023effective,garibi2024renoise} update the inversion noise at each timestep, but requiring excessive time. (b) \emph{Ordinary differential equation} (ODE) based approaches~\cite{rf-inv,rf-solver} design invertible sampler with better efficiency, but are often at the cost of reconstruction quality. (c) DeepInv is the first approach to train an step-by-step inversion solver to fast and accurately predict inversion noise.}
  \label{fig:methods}
\end{figure}

\begin{figure*}[t]
  \centering
  \includegraphics[width=1\linewidth]{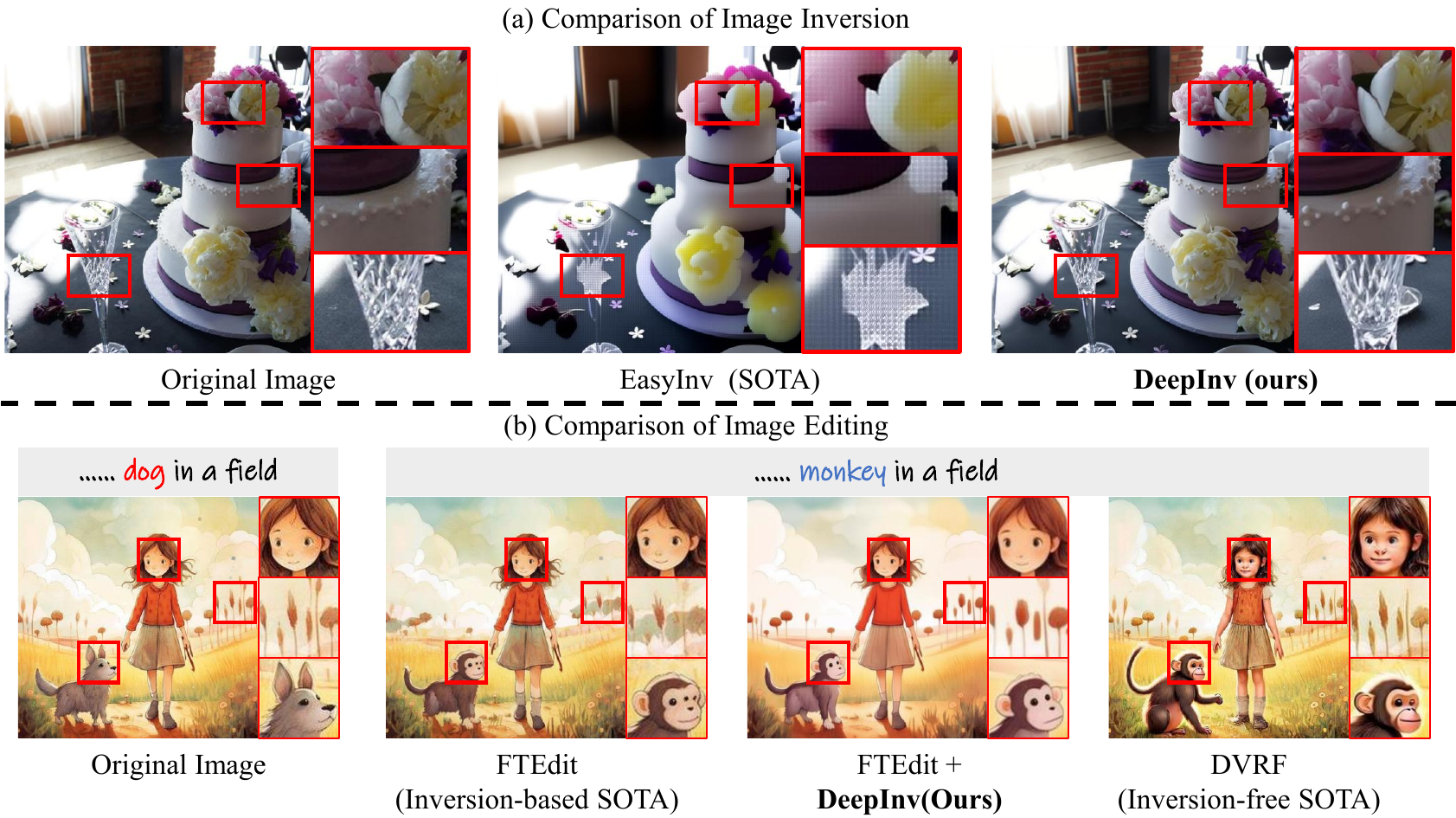}
  \caption{The comparison between our DeepInv and existing SOTA methods, \emph{i.e.}, EasyInv~\cite{EasyInv}, FTEdit~\cite{FTEdit} and DVRF~\cite{DVRF}, in terms of image inversion (a) and editing (b). (a) shows the visualization of image inversions, which shows that our DeepInv can better preserve the details of the original images than EasyInv, \emph{e.g.}, the textures and structure. (b) shows the image editing results according to the text prompt. The inverted noise predict by DeepInv can help FTEdit achieve more better editing results well aligned to the text prompt, while the compared methods are easy to fail in alignments.}
  \label{fig:Large_compare}
\end{figure*}

%% file: sec/1_intro.tex
\section{Introduction}
\label{sec:intro}

Recent years has witnessed the great breakthroughs made by image diffusion models in the field of image generation~\cite{hu2024anomalydiffusion,liu2024adv,ji2024diffusion,nguyen2025hedit}. Representative diffusion models, such as \emph{Stable Diffusion} series~\cite{rombach2022high,podell2023sdxl,esser2024scaling} and \emph{DALL-E3}~\cite{betker2023improving}, exhibit much superior capabilities than previous  generative approaches~\cite{goodfellow2014generative,reed2016generative,li2022interpretable}. 
In terms of diffusion-based image editing, a critical application of diffusion models \cite{wang2024compositional,mou2024dragondiffusion}, \emph{diffusion inversion} is a research hot-spot that attracts increasing attention from both academia and industry \cite{pan2023effective,garibi2024renoise,rf-solver,rf-inv,Nguyen_2025_CVPR}.
This task aims to achieve a mapping between the noise space and the real images, based on which the diffusion models can achieve accurate and controllable image generation or manipulation~\cite{hertz2022prompt,cao_2023_masactrl,gao2025postermaker}. Despite the progress, existing diffusion inversion methods are still hard to achieve satisfactory results in both performance and efficiency simultaneously due to lack of supervisions.
In particular, a key challenge of diffusion inversion is the absence of ground-truth annotations~\cite{pan2023effective}, which are intractable to obtain. In this case, conventional supervised learning paradigms for image inversion are hard to implement, and the researchers have to shift towards approximation-based strategies, such as the \emph{iterative optimization}~\cite{garibi2024renoise,pan2023effective} and \emph{ordinary differential equation} (ODE) based ones~\cite{wallace2023edict,rf-inv,rf-solver}, as shown in Fig.~\ref{fig:methods}.
While these efforts have achieved remarkable progresses, they still encounter several key issues, such as computation complexity~\cite{garibi2024renoise}, inversion instability~\cite{wallace2023edict,rf-inv,rf-solver} or low reconstruction quality~\cite{cao_2023_masactrl}. As shown in Fig.\ref{fig:Large_compare}, although the SOTA method EasyInv \cite{EasyInv} can greatly shorten the reconstruction process, it is still hard to handle the images with complex textures or structure details. Besides, on downstream tasks, inversion-free approach DVRF~\cite{DVRF} results in inconsistency across non-editing area, and FTEdit~\cite{FTEdit}, as a inversion-based method, generate a money with four eyes. Overall, achieving the fast, stable and high-quality diffusion inversion still remains an open problem.

To overcome these challenges we propose a novel self-supervised diffusion inversion  approach in this paper, termed \emph{Deep Inversion} (DeepInv).
The main principle of DeepInv is to explore effective self-supervision signals to directly train an parameterized solver, thereby achieving efficient and effective diffusion inversion. To approach this target, DeepInv first resort to \emph{fixed-point iteration theory}~\cite{anderson1965iterative} to automatically derives pseudo noise annotations from real images, of which procedures requires no manual intervention. Besides, a novel data augmentation strategy based on linear interpolation is also proposed to fully exploit the pseudo information from limited real images, and these high-quality pseudo labels are used as the self-supervision signals.
Based on the above designs, DeepInv further introduce a novel training regime to integrate pseudo label generation and self-supervised learning in one unified framework, where a multi-scale learning principle is also adopted to progressively improves the solver's capability for diffusion inversion.
With these innovative designs, DeepInv can effectively train a parameterized diffusion solver to directly predict the noise of the given images, improving the efficiency and performance by orders of magnitudes. Moreover, we also carefully design two trainable solver networks for SD3~\cite{esser2024scaling} and Flux~\cite{labs2025flux1kontextflowmatching}, respectively, which can also enlighten the research fo community.

To validate DeepInv, we conduct extensive experiments on the COCO~\cite{EasyInv} and PIE-Bench~\cite{ju2024pnp} benchmarks. The experimental results show the comprehensive improvements of DeepInv than existing methods in terms of inversion efficiency and quality, \emph{e.g.}, +200\% SSIM than ReNoise~\cite{garibi2024renoise} on COCO dataset~\cite{COCO} with +9887.5\% faster speed. Besides, in-depth analyses as well as downstream task performance further confirm the merits of DeepInv.

\hspace*{\fill} \\

\noindent Overall, our contributions are two-fold:

\begin{itemize}
    \item We present the first self-supervised based approach for diffusion inversion, termed DeepInv, which can help to train parameterized inversion solvers for the fast and accurate mapping of inversion noises.
    \item Based on DeepInv, we propose the first two parameterized inversion solver for SD3.5 and FLUX, respectively, yielding comprehensive improvements than existing inversion methods while providing insights to community.
\end{itemize}

%% file: sec/2_related.tex
\begin{figure*}[!t]
  \centering
  \includegraphics[width=1\linewidth]{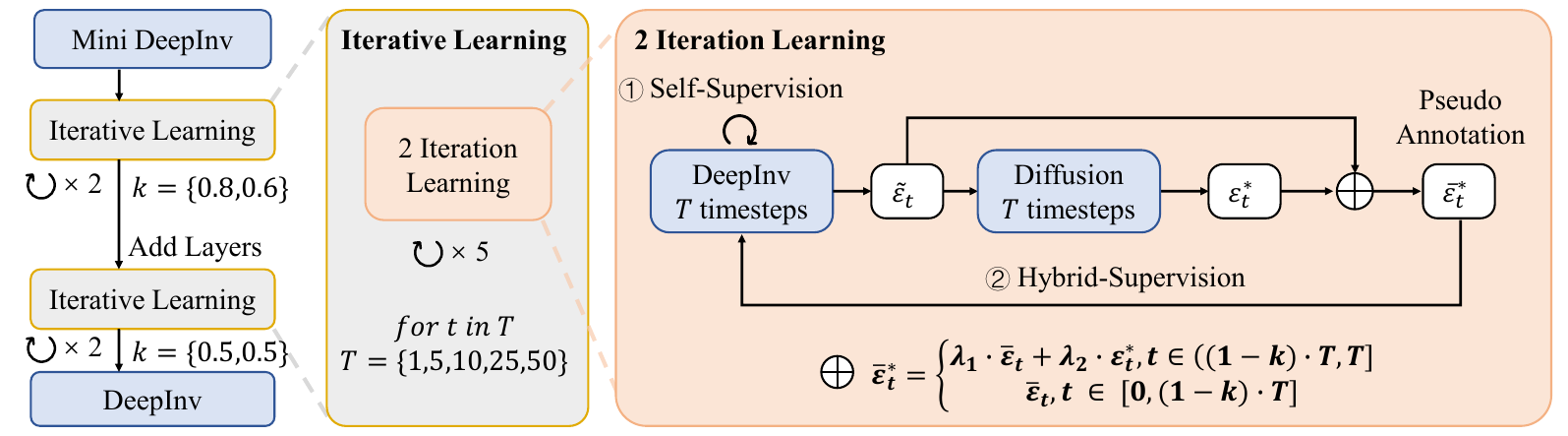}
  \caption{{Overview of our proposed training framework.} We begin by initializing a base network of the inversion solver. The training process proceeds through 4 iterative stages. In each iteration, the solver is trained under 5 different timestep configurations, and for each configuration, 2 rounds of optimization are performed using distinct loss functions. 
  During the \engordnumber{3} iteration, additional layers are appended and trained while the previously learned parameters are frozen. In the \engordnumber{4} iteration (the last one), all parameters are jointly fine-tuned with a learning rate reduced to 10\% of the original. Further implementation details are provided in Section~\textbf{Method}.} 
  \label{fig:Ours_framework}
\end{figure*}

\section{Related Works}
\label{sec:related}

Diffusion models advance generative modeling by iteratively reversing the noise corruption process. DDPM~\cite{ho2020denoising} formalizes this with stable training but incurs high sampling costs. DDIM~\cite{song2020denoising} later introduces a deterministic formulation that reduces sampling steps while maintaining quality. Recently, a series of novel designs have introduced the next generation of diffusion models~\cite{esser2024scaling,flux2024,labs2025flux1kontextflowmatching}, leading to further improvements in performance. Another key innovation in diffusion models is the introduction of rectified flow~\cite{liu2023flow}, which proposes a novel training paradigm to align the denoising directions across timesteps. By encouraging consistent noise prediction trajectories, this approach enables diffusion models to perform inference with fewer steps while maintaining output quality, thereby improving efficiency. However, these changes introduce new challenges on the need for dedicated inversion methods specifically tailored to rectified-flow-based models, which enable efficient and high-fidelity generation in modern diffusion architectures.

In particular, DDIM Inversion~\cite{couairon2023diffedit} represents one of the first attempts by introducing a reverse denoising mechanism to recover the noise of given images while preserving structural coherence. Null-Text Inversion~\cite{mokady2023null} enhances reconstruction quality by decoupling conditional and unconditional textual embeddings, suggesting that empty prompts can have a positive impact on the inversion task. Another approach, PTI~\cite{dong2023prompt}, refines prompt embeddings across denoising steps to achieve better alignment. ReNoise~\cite{garibi2024renoise} applies cyclic optimization to iteratively refine noise estimates, while DirectInv~\cite{ju2023direct} introduces latent trajectory regularization to mitigate inversion drift. More recently, Rectified Flow~\cite{liu2023flow} reshapes the inversion landscape by enforcing consistent noise trajectories across timesteps, achieving high-quality generation with significantly reduced inference costs. This technique is adopted by advanced architectures such as Flux~\cite{flux2024} and SD3~\cite{esser2024scaling}, prompting the development of new inversion methods. RF-Inversion~\cite{rf-inv} and RF-Solver~\cite{rf-solver} both focus on solving the rectified-flow ODE for inverting the denoising process, each proposing distinct solutions within this framework. Text-guided editing methods align textual and visual features for controlled generation. Prompt-to-Prompt~\cite{hertz2023prompt} manipulates attention maps for attribute edits but requires full denoising. Later works such as Plug-and-Play~\cite{tumanyan2023plug} and TurboEdit~\cite{chen2024turboedit} improve efficiency through latent blending and time-shift calibration, with TurboEdit~\cite{chen2024turboedit} also introducing guidance scaling to enhance edit strength without compromising background consistency. However, many of these methods are not publicly available or require extensive training resources. In contrast, our method provides a lightweight and open-source alternative with competitive performance and efficiency.

%% file: sec/3_pre.tex
\section{Preliminary}
\label{sec:foundation}
 Currently, diffusion inversion is often regarded as a \emph{fixed-point problem} in existing research~\cite{pan2023effective,garibi2024renoise}. For a given function $f(x)$, the general fixed-point equation can is
\begin{equation}\label{eq:fixed}
x_0 = f(x_0),
\end{equation}
where \(x_0\) denotes the fixed point of $f(x)$. For diffusion inversion, we aim to find a mapping \(g(\cdot)\) that transforms the latent codes between consecutive timesteps $t$
\begin{equation}\label{eq:inv}
z_{t+1} = g(z_t).
\end{equation}
Given an optimal latent code \(\bar z_t\), its inversion \(\bar{z}_{t+1} = g(\bar{z})\) obviously should satisfy the add-noise-denoising consistency condition~\cite{EasyInv}, which means a good-quality inversion noise should be consistent with the denosing ones at the same time step:
\begin{equation}\label{eq:deno}
\bar{z}_{t} = d(g(\bar{z}_{t})),
\end{equation}
where \(d\) represents the denoising process.

Let \(F = d \circ g\) denote the composite function, then Eq.\ref{eq:deno} can be transformed into a fixed-point form:
\begin{equation}\label{eq:fixed_inv}
\bar{z}_{t} = F(\bar{z}_{t}).
\end{equation}
Here, Eq.\ref{eq:fixed_inv} is the principle rule we use for inversion tasks, laying the theoretical base for DeepInv.

\begin{figure}[t]
  \centering
  \includegraphics[width=1\linewidth]{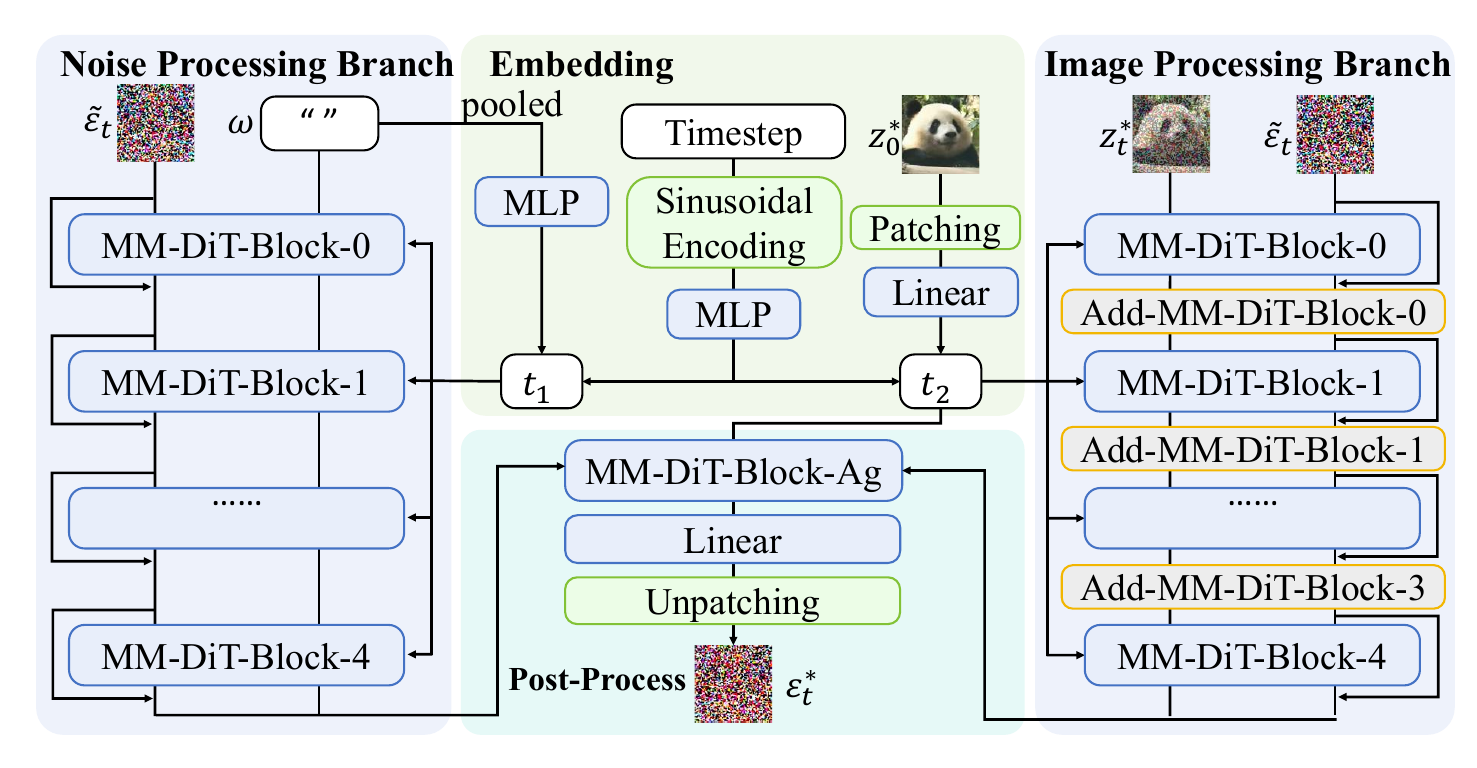}
  \caption{Architecture of the proposed DeepInv solver. The dual-branch design processes noise (left) and image (right) information separately before final aggregation. Extra blocks added in second last round to extend model ability.}
  \label{fig:Ours_Struc}
\end{figure}

%% file: sec/4_method.tex
\section{Method}
\subsection{DeepInv}

In this paper, we present a novel self-supervised learning regime for diffusion inversion, termed \textbf{DeepInv}, as depicted in Fig.~\ref{fig:Ours_framework}. To enable the effective training of parameterized inversion solvers, DeepInv introduces a novel self-supervised learning design, considering the temporal consistency, progressive refinement and multi-scale training for diffusion inversion. 

\subsubsection{Overview}
DeepInv operates on latent representations derived from a pretrained diffusion model. Given given an input image \(I\), we first encodes it into a latent code \(z_0^*\) through diffusion model's original VAE:
\begin{equation}\label{eq:vae}
z_0^* = \mathrm{VAE}(I).
\end{equation}
Starting from $z_0^*$, the parameterized solver $g^*(\cdot)$ predicts the inverse noise step by step:
\begin{equation}\label{eq:in-out}
g^*(z_t^*, t) = \varepsilon^*_t,
\end{equation}
where $\varepsilon^*_t$ denotes the predicted inversion noise. It is added to latent $z_t^*$ by diffusion model's original sampler $s$:
\begin{equation}\label{eq:sampler}
s(z_t^*, -\varepsilon^*_t) = z_{t+1}^*.
\end{equation}
Here, we consider $-\varepsilon^*_t$ as the opposite noise, since diffusion inversion is a process of adding noise. 

Thus, the solver outputs temporally consistent estimations that can reconstruct the underlying clean latent $z_0^*$.
In this case, the overall objective of DeepInv is to enable correctly and temporally coherent inversion across arbitrary diffusion models. Formally, the solver seeks to minimize the discrepancy between predicted noise/latents and their counterparts derived from the forward process, thereby approximating the true inverse dynamics. This objective can be formulated as
\begin{equation}\label{eq:obj_inversion}
\underset{\phi}{min}~\mathbb{E}\Big[\Vert g_{\phi}^*(z_t^*, t) - d^*(z_{t+1}^*, t+1)\Vert_2^2\Big].
\end{equation}

As we mentioned, when inverted noise is exactly same as the denoised one in the same timestep, it would be an ideal inversion, that is what this objective function targeting at. Considering Eq.\ref{eq:obj_inversion} as the self-supervised objective, the corresponding loss function of DeepInv is defined by 
\begin{equation}\label{eq:l2_loss}
\begin{split}
    \mathcal{L}_{self} &= \Vert\varepsilon_t^*-\bar\varepsilon_t\Vert_2^2.
\end{split}
\end{equation}
In particular, the reference noise $\bar\varepsilon_t$ is obtained by
\begin{equation}\label{eq:denoising}
\bar\varepsilon_t = d^*(z^*_{t+1},t+1),
\end{equation}
where \(d^*\) represents the pretrained diffusion model, and \(z^*_{t+1}\) is generated by adding noise \(\varepsilon_t^*\) to \(z^*_t\). With Eq.\ref{eq:l2_loss}, our DeepInv can perform the parameterized solver's optimization without ground truth.

Besides, a hybrid loss is also used to incorporate pseudo labels generated from denoising-inversion fusion:
\begin{equation}\label{eq:loss_mx}
\mathcal{L}_{hyb} = \Vert\varepsilon_t^*-\bar{\varepsilon}_t^*\Vert_2^2,
\end{equation}
where $\bar{\varepsilon}_t^*$ represents the fused pseudo noise, which will be detailed later. During multi-scale fine-tuning stage, DeepInv employs a stabilized loss function defined by
\begin{equation}\label{eq:loss_new}
\mathcal{L}_{stable} = \alpha\cdot\mathcal{L}_{self} + (1 - \alpha)\cdot\mathcal{L}_{hyb},
\end{equation}
where $\alpha$ is the hyper-parameter to balance the smooth of optimization and over-fitting.

\begin{table*}[!t]
\caption{Comparison between our DeepInv Solver and existing diffusion inversion methods on COCO~\cite{COCO}. Our DeepInv Solver can achieve obvious performance gains while retaining high efficiency.}
\centering
\small
 {\begin{tabular}{c | c c c c c c} 
 \toprule
 & LPIPS ($\downarrow$) & SSIM ($\uparrow$) & PSNR ($\uparrow$) & MSE ($\downarrow$)& FID ($\downarrow$) & Time ($\downarrow$) \\ [0.5ex] 
 \midrule
 DDIM Inversion~\cite{couairon2023diffedit} & 0.452 & 0.501 & 12.936 & 0.059 & 187.528 & 34\textbf{s}\\
 \hline
 RF Inversion~\cite{rf-inv} & 0.380 & 0.507 & 16.105 & 0.027 & 180.372 & 34\textbf{s}\\
 \hline
 ReNoise~\cite{garibi2024renoise} & 0.614  & 0.451  & 12.152  & 0.071 & 250.882 & 4746\textbf{s}\\
 \hline
 EasyInv~\cite{EasyInv} & \underline{0.294} & \underline{0.643} & \underline{18.576} & \underline{0.021} & \underline{153.333} & {34}\textbf{s}\\
 \hline
 DeepInv Solver~(\textbf{Ours}) & \textbf{0.075} & \textbf{0.903} & \textbf{29.634} & \textbf{0.001} & \textbf{37.879} & 48\textbf{s}\\
 \bottomrule
\end{tabular}}
\label{precision_table}
\end{table*}

\subsubsection{Pseudo Label Generation}
To further enhance inversion quality, we adopt a pseudo annotation strategy inspired by \emph{Null-Text Inversion}~\cite{mokady2023null}. Instead of optimizing text embeddings via gradient descent, we directly fuse the denoising and inversion noise predictions, formulated as
\begin{equation}\label{eq:add_noise}
\bar{\varepsilon}_t^* = 
\begin{cases}
\lambda_1 \cdot\bar{\varepsilon}_t + \lambda_2\cdot\varepsilon_t^*, & t \in ((1 - k)\cdot T,T] \\[4pt]
\bar{\varepsilon}_t. & t \in [0, (1 - k)\cdot T]
\end{cases}
\end{equation}
Here, $\bar{\varepsilon}_t$ denotes the noise predicted by the diffusion model, which serves as the theoretical object for our inversion solver. However, in practice, $\bar{\varepsilon}_t$ tends to accumulate errors during the denoising process, leading to inaccurate estimations. To address this issue, we incorporate the inversion noise $\varepsilon^*_t$ generated by our solver $g^*(\cdot)$, combining it with $\bar{\varepsilon}_t$ for more stable learning. As demonstrated in \emph{Null-text inversion}~\cite{mokady2023null}, aligning the predicted denoising noise with the inversion noise effectively enhances inversion accuracy and reconstruction fidelity. The coefficients $\lambda_1$ and $\lambda_2$ are introduced to balance the relative contributions of these two noise components in the optimization objective. This linear fusion captures both high-confidence denoising outputs and transitional inversion predictions, based on which we further construct a synthesized dataset $\mathcal{Q}$ that provides stable and high-quality pseudo supervision. Compared to gradient-based optimization, this mechanism achieves equivalent alignment quality with significantly improved computational efficiency.

\subsubsection{Training Procedure}
In terms of training process, DeepInv adopts an iterative and multi-scale training regime to progressively refine inversion accuracy and temporal coherence, as depicted in Fig. \ref{fig:Ours_framework}.

\textbf{Iterative Learning.}  
The training process is divided into multiple temporal stages $\mathcal{T} = \{1, 5, 10, 25, 50\}$, where the numbers denote the timestep for each round of training. Each stage captures a distinct temporal resolution, considering the diffusion process as a \emph{vector field}~\cite{esser2024scaling}. In DeepInv, low resolutions training serves to capture global trajectory patterns, while the higher ones focus on local details. The training begins with a low-resolution warm-up ($\mathcal{T}_0 = 1$), and proceeds by progressively increasing temporal resolution, with parameters inherited between stages. The final stage ($\mathcal{T}_4 = 50$) aims to obtain full temporal coherence. In practice, $\mathcal{T}$ could be changed, \emph{e.g.}, FLUX-Kontext~\cite{labs2025flux1kontextflowmatching} designs to generate image in fewer steps, while each of them takes long time. In this case, $\mathcal{T}$ could be set to $\{1, 5, 10\}$.

\textbf{Multi-scale Tuning.}  
To balance efficiency and capacity, the model depth scales with the value of $k$. A 5-layer configuration is used for $k \in [0.8, 0.6]$, and the depth increases to 9 layers when $k=0.5$. As illustrated in Fig.~\ref{fig:Ours_Struc}, new layers are appended to the right branch with residual connections for stability.  
Fine-tuning proceeds in two iterations:  
(1) only the newly added layers are optimized, while others remain frozen;  
(2) all parameters are fine-tuned with a reduced learning rate (10\% of the original).  
This progressive scheme allows DeepInv to preserve previously learned inversion knowledge while continuously improving reconstruction fidelity across scales. 

\subsection{DeepInv Solver}
Based on DeepInv, we further propose innovative parameterized solvers for existing diffusion model~\cite{esser2024scaling}, which aims to exploit the pre-trained diffusion knowledge and image-specific cues through a dual branch architecture. Here, we use \emph{Stable Diffusion 3} \cite{esser2024scaling} as the base model, showing the construction of DeepInv's solver. The other solver for Flux are depicted in appendix due to the page limit, which also follows the same principle. Concretely, \emph{DeepInv Solver} is built based on the same components of SD3 with additional refinement modules for noise correction, as shown in Fig.~\ref{fig:Ours_Struc}. This design maintains the information of DDIM inversion noise~\cite{couairon2023diffedit}, while improving accuracy through residuals.

\textbf{Input of DeepInv Solver.}  
The diffusion model takes the latent $z_t^*$, timestep $t$, and prompt $\omega$ as inputs to predict the corresponding noise. For our inversion solver $g^*(\cdot)$, to achieve higher reconstruction fidelity, we additionally introduce the DDIM inversion noise $\tilde{\varepsilon}_t$ as an auxiliary input:
\begin{equation}
\tilde{\varepsilon}_t = \tilde{d}(z_t^*, t),
\end{equation}
where \(\tilde{d}\) denotes the DDIM inversion operator, and \(z_t^*\) represents the denoised latent at timestep \(t\).  
By incorporating DDIM inversion noise $\tilde{\varepsilon}_t$, which serves as a well-established baseline in inversion, the solver gains a strong initialization prior. Moreover, residual connections are applied between the output of our solver and $\tilde{\varepsilon}_t$ as well as the modules, ensuring that the final performance will not fall below the baseline.

\textbf{Dual-Branch Design.}  
A key innovation of the DeepInv Solver lies in its \textit{dual-branch architecture}, which separates pretrained prior modeling (left branch) from image-conditioned refinement (right branch). This structural disentanglement enables a balanced trade-off between inversion fidelity and structural consistency, allowing the model to achieve high-quality reconstruction without requiring explicit ground-truth noise supervision.  

Among the four types of inputs to our solver, the latent $z_t^*$ is fed into the right branch to facilitate image-guided refinement, complementary to the left branch, as shown in Fig. \ref{fig:Ours_Struc}. The prompt embedding $\omega$ is sent to the left branch but is typically set to an empty token, following the strategy of \emph{Null-Text Inversion}~\cite{mokady2023null}, which shows that empty prompt conditioning enhances inversion fidelity. Meanwhile, the DDIM inversion noise $\tilde{\varepsilon}_t$ serves as a shared input to both branches, providing a high quality prior across the model.

Another novel design is the use of two distinct timestep embeddings, reflecting their respective branch functionalities. The left-branch embedding \(t_1\) follows the original SD3 temporal encoding, constructed from SD3’s temporal embedding module together with \(\omega\). It also keeps the full compatibility with the pretrained model, as shown in Fig.~\ref{fig:Ours_Struc}. For the right branch, the timestep embedding \(t_2\) is defined by
\begin{equation}
t_2 = \mathrm{TEMB}(z_0^*, t),
\end{equation}
where \(z_0^*\) is used in place of the prompt embedding to preserve visual coherence and retain original image information during the inversion process. Both branches consist of stacked MM-DiT blocks~\cite{Peebles2023DiT,esser2024scaling}, and the conditional vectors $t_1$ and $t_2$ remain consistent in their respective branches. Residual connections are employed both between the blocks and at the final output, which can retain the prior information from DDIM inversion, and progressive refinement through layers. After feature extraction, the two branches are fused through an MM-DiT aggregation block followed by a linear projection layer to generate the predicted noise. Finally, residual addition is applied between the predicted noise and $\tilde{\varepsilon}_t$, yielding the final output. Details of the Flux solver can refer to appendix.

%% file: sec/5_exp.tex
\begin{table*}[!t]
\caption{Comparison of image editing task on the PIE~\cite{ju2024pnp} benchmark. DeepInv Solver are combined with two diffusion editing methods to show the benefit of better inversion noises, \emph{i.e.}, FTEdit~\cite{FTEdit} and RF Inversion~\cite{rf-inv}. DVRF is the SOTA and inversion-free method.}
\centering
\small
 {\begin{tabular}{c | c c c c c c} 
 \toprule
 & LPIPS ($\downarrow$) & SSIM ($\uparrow$) & PSNR ($\uparrow$) & MSE ($\downarrow$)& FID ($\downarrow$) \\ [0.5ex] 
 \midrule
 FTEdit~\cite{FTEdit} & {\textbf{0.078}} & {\textbf{0.90}} & \underline{25.117} & \underline{0.004} & \underline{44.423}\\ 
 \hline
 FTEdit + DeepInv Solver & \underline{0.087} & {\textbf{0.90}} & {\textbf{26.255}} & {\textbf{0.003}} & {\textbf{41.158}}\\ 
 \hline
 \hline
 RF Inversion~\cite{rf-inv} & 0.211 & 0.71 & 19.855 & 0.014 & 67.787\\ 
 \hline
 RF Inversion + DeepInv Solver & {0.111} & {0.86} & {24.519} & {0.005} & 54.056 \\ 
 \hline
 \hline
 DVRF~\cite{DVRF} & 0.093  & 0.85  & 23.372  & 0.007 & 56.698\\ 
 \bottomrule
\end{tabular}}
\label{downstream}
\end{table*}

\begin{table}[t]
\caption{The impact of noise interpolation strategy, for different inversion methods. With same strategy, our approach performed the best. The base model used is SD3.}
\centering
\large
\resizebox{1\columnwidth}{!}
 {\begin{tabular}{c | c || c c c c c} 
 \toprule
 +Noise Interpolation & $k$ & LPIPS ($\downarrow$) & SSIM ($\uparrow$) & PSNR ($\uparrow$) & MSE ($\downarrow$)& FID ($\downarrow$) \\ [0.5ex] 
 \midrule
 DDIM Inversion & 0.5 & 0.245 & 0.785 & 23.347 & 0.005 & 108.718 \\
 \hline
 EasyInv & 0.5 & 0.192 & 0.745 & 25.042 & 0.004 & 103.709 \\
 \hline
 DeepInv Solver~(\textbf{Ours}) & 0.5 & \textbf{0.075} & \textbf{0.903} & \textbf{29.634} & \textbf{0.001} & \textbf{37.879} \\
 \bottomrule
\end{tabular}}
\label{add_noise_compare}
\end{table}

\begin{table}[t]
\caption{Ablation study of the number of adding layers to DeepInv Solver. The base model used is SD3.}
\centering
\resizebox{1\columnwidth}{!}
 {\begin{tabular}{ c | c || c c c c c} 
 \toprule
  Layers & Branch & LPIPS ($\downarrow$) & SSIM ($\uparrow$) & PSNR ($\uparrow$) & MSE ($\downarrow$)& FID ($\downarrow$) \\ [0.5ex] 
 \midrule
  5 & None & 0.076 & 0.900 & 28.652 & 0.002 & 38.106 \\
 \hline
  9 & Both & 0.076 & \textbf{0.903} & 29.563 & \textbf{0.001}  & 38.188  \\
 \hline
  9 & Right & \textbf{0.075} & \textbf{0.903} & \textbf{29.634} & \textbf{0.001} & \textbf{37.879} \\
 \bottomrule
\end{tabular}}
\label{ablation_exp}
\end{table}

\section{Experiments}
\label{sec:exp}

We validate the effectiveness of our method through comprehensive experiments on the COCO \cite{COCO} and PIE-Bench \cite{ju2024pnp} benchmarks, and compare it with a set of advanced diffusion inversion methods as well as the baseline method, including DDIM inversion~\cite{couairon2023diffedit}, EasyInv~\cite{EasyInv}, ReNoise~\cite{garibi2024renoise} and RF-Inversion~\cite{rf-inv}. 

\subsection{Experimental Settings}
\label{sec:exp_set}
\subsubsection{Implementation details }
To ensure a fair comparison, each inversion method is first applied to estimate the noise corresponding to input images. The resulting noise is then fed into the SD3 model to reconstruct the images, which are subsequently used for evaluation. We run each method for one time. Experiments are conduct on three NVIDIA GeForce RTX 3090 GPUs, with 24GB usable memory each. For some of these methods are not public or does not support SD3 inversion, we re-implement them for following experiments.
For hyper-parameters, we have $\lambda_1=\lambda_2=0.5$, and $k\in [0.8,0.6,0.5,0.5]$. Our batch size is set to 4 due to the limitation of GPU's saving space. We also have a dynamical training epoch setting, $epoch \in [300,300,250,200,100]$, corresponding to the temporal stages $\mathcal{T}$. The feature dimension of MM-DiT blocks of our solver are same as their original setting in SD3 model.

\subsubsection{Datasets and metrics}
We use two mainstream benchmarks. COCO~\cite{COCO} is a large-scale image collection originally created for object detection, segmentation and captioning: it contains over 300,000 images across 80 object categories and more than 2.5 million labeled instances. The PIE-Bench~\cite{ju2024pnp} is a prompt-driven image editing dataset comprising 700 images paired with editing instructions across diverse scene types and editing operations (e.g., object addition, removal, attribute modification) designed to evaluate text-driven image editing performance.
In terms of the self-supervised training of DeepInv, we also create a dataset of real images from the COCO dataset~\cite{COCO}, selecting samples with near-square aspect ratios to ensure compatibility with all inversion framework. The selected images span a wide range of categories, including animals, objects, and other diverse content, and are resized to a resolution of $1024 \times 1024$ pixels. The dataset is split into 2,000 images for training and 298 images for testing, with no overlap between the two sets.
Following previous works~\cite{mokady2023null,garibi2024renoise,ju2024pnp}, we adopt the widely used inversion metrics, including LPIPS~\cite{zhang2018unreasonable}, SSIM~\cite{wang2004image}, PSNR~\cite{hore2010image}, MSE and FID~\cite{heusel2017gans} as evaluation metrics. 


\subsection{Quantitative Results}
\label{sec:quanti}
\textbf{Comparison with existing methods.}  
We first compare our DeepInv with several state-of-the-art (SOTA) inversion methods in Tab.~\ref{precision_table}. From this table, we can first observe that traditional inversion methods such as DDIM Inversion~\cite{couairon2023diffedit} and EasyInv~\cite{EasyInv} achieve limited reconstruction fidelity, as reflected by the relatively low PSNR and SSIM values. For instance, DDIM Inversion shows significant degradation under accumulated errors, while EasyInv improves upon DDIM but still struggles to maintain fine-grained visual consistency. This can be attributed to their reliance on either handcrafted inversion trajectories or simple noise estimation without self-supervised refinement. In contrast, our DeepInv achieves substantial improvements across all metrics, with a remarkable +129.1\% gain in PSNR and +80.2\% in SSIM compared to DDIM Inversion, and clear advantages over EasyInv (+75.3\% in FID and +74.5\% in LPIPS). Moreover, DeepInv demonstrates superior efficiency, \emph{e.g.}, operating 9887.5\% faster than the iterative ReNoise~\cite{garibi2024renoise} method, while requiring only 14 additional seconds compare to DDIM Inversion, which represents the theoretical upper limit of inversion task. These experiments well confirm the effectiveness of DeepInv in achieving high-fidelity and high-efficiency diffusion inversion.

\textbf{Comparison on downstream editing tasks.}  
To further evaluate the versatility of DeepInv, we integrate it into two representative diffusion-based editing approaches, \emph{i.e.}, RF Inversion~\cite{rf-inv} and FTEdit~\cite{FTEdit}, and report the results in Tab.~\ref{downstream}. We also include DVRF~\cite{DVRF}, a leading inversion-free editing model, as a reference baseline for end-to-end diffusion editing. The comparison reveals several interesting observations. Firstly, without our solver, inversion-based methods such as RF Inversion and FTEdit often struggle to maintain both structural fidelity and visual coherence, showing inconsistent texture reconstruction and noticeable semantic drift. For example, RF Inversion tends to significant inconsistency on local details though preserving global layout, while FTEdit occasionally suffers from unnatural transitions around edited areas. These artifacts mainly arise from imperfect or unstable inversion noise estimation, which directly affects the downstream editing quality. In stark contrast, when applying our DeepInv solver, both RF Inversion and FTEdit demonstrate substantial and consistent improvements across nearly all metrics. DeepInv solver not only stabilizes the inversion process but also enhances semantic alignment and texture consistency, allowing these methods to outperform the inversion-free DVRF baseline on multiple quantitative indicators. The only exception appears in FTEdit’s LPIPS score, in which the original score is slightly higher. Overall, these results highlight that DeepInv serves as a powerful and general inversion backbone, \emph{i.e.}, being capable of elevating diverse diffusion-based editing pipelines by providing more faithful inversions. These results also reveal that inversion-based editing frameworks possess significant potential limitations in their prior performance, which may primarily stem from the lack of a proper inversion mechanism rather than the editing formulation itself.

\begin{figure*}[!t]
  \centering
  \includegraphics[width=1\linewidth]{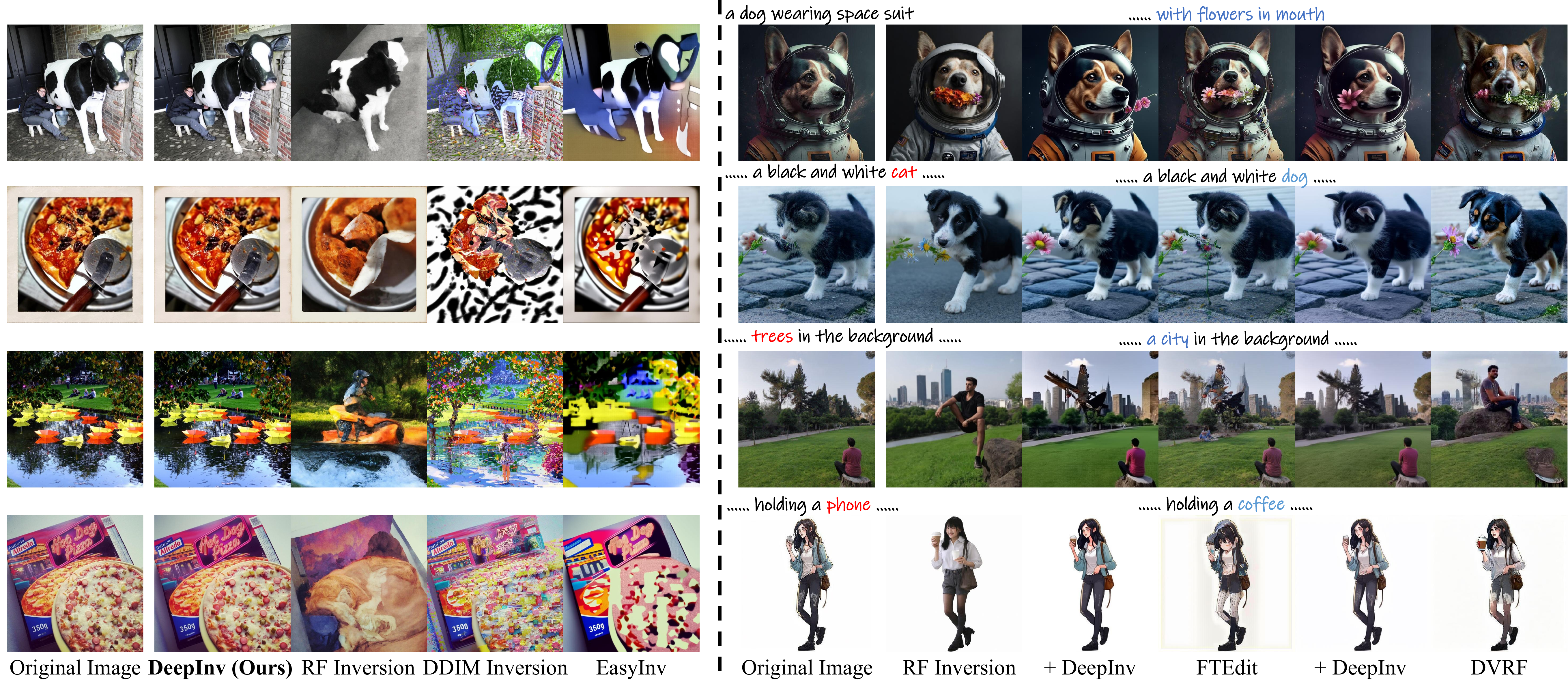}
  \caption{Visualized comparison between DeepInv Solver and existing methods on the task of image inversion (left) and editing (right), respectively. For the image editing task, we integrate DeepInv solver into two representative inversion-based diffusion editing methods, \emph{i.e.}, RF-Inv~\cite{rf-inv} and FTEdit~\cite{FTEdit}, by replacing their original inversion modules. DVRF~\cite{DVRF} is a SOTA and inversion-free method. In each example, the object in the original image is marked in {\color{red}red}, and the replaced (or added) object is highlighted in {\color{blue}blue}. The first-row example illustrates an object-addition scenario with only the {\color{blue}blue} prompts. According to shown images, DeepInv solver consistently achieves more faithful inversions and leads to visually coherent edits.}
  \label{fig:visual_compare_downstream}
\end{figure*}

\textbf{Ablation Studies.} We then ablate the key designs of DeepInv in Tab.\ref{add_noise_compare} and Tab.\ref{ablation_exp} to examine the impact of our noise interpolation module , \emph{i.e.}, Eq.\ref{eq:add_noise}. We conduct a controlled experiment by applying the same interpolation strategy to two strong baselines, \emph{i.e.}, DDIM inversion and EasyInv. As shown in Tab.\ref{add_noise_compare}, even under this setup, our method consistently outperforms both baselines by a large margin. This result indicates that while noise interpolation contributes positively, the gains achieved by our framework does not solely rely on this component, but instead reflect the overall effectiveness and robustness of our design. Tab.\ref{ablation_exp} summarizes the impact of layer extension. We observe that additional layers generally improves performance, confirming the benefit of increased model capacity. However, adding layers to both branches leads to degraded results and increased computational cost. We attribute this to the processing of DDIM-inverted noise in the left branch which provides high quality prior information. It requires minimal modification, and excessive complexity may hinder convergence and introduce instability. Thus, the configuration that adds layers only to the right branch proves most effective and become our final choice.

\subsection{Qualitative Results}
\label{sec:qualitative}
To further demonstrate the effectiveness of the proposed DeepInv, we visualize its performance on a range of downstream image editing and inversion tasks in Fig.~\ref{fig:visual_compare_downstream}, which includes comparisons with both diffusion-based editing pipelines and recent advanced inversion approaches.

In the left part of Fig.~\ref{fig:visual_compare_downstream}, we present a comparison between DeepInv and other advanced inversion methods, including DDIM Inversion~\cite{couairon2023diffedit}, EasyInv~\cite{EasyInv}, and ReNoise~\cite{garibi2024renoise}. While DDIM Inversion and EasyInv can produce structurally coherent outputs, they often fail to preserve high-frequency visual details.  Although being capable of reconstructing global layouts, ReNoise~\cite{garibi2024renoise} tends to introduce color inconsistencies and unnatural tones. In contrast, DeepInv yields reconstructions that remain faithful to the original images in both structure and texture, achieving nearly imperceptible differences between the reconstructed and original visuals. These results confirm that DeepInv provides the most faithful and perceptually realistic inversions among existing methods. In the right part of Fig.~\ref{fig:visual_compare_downstream}, we evaluate DeepInv on downstream editing tasks by integrating it into two representative inversion-based diffusion editing methods namely RF-Inv~\cite{rf-inv} and FTEdit~\cite{FTEdit}. We replacing their original inversion modules by our DeepInv solver. We also compare against DVRF~\cite{DVRF}, a SOTA and inversion-free editing method. As shown, the use of DeepInv markedly improves reconstruction fidelity and semantic consistency across both edited and non-edited regions further confirming the contributions of our work. Compared to the inversion-free DVRF, methods equipped with our solver better preserve fine-grained details and maintain stronger spatial alignment between the edited object and the original background. These qualitative comparisons highlight that DeepInv not only strengthens inversion-based editing pipelines but also ensures superior structural coherence and visual realism across diverse editing scenarios.

%% file: sec/6_conclu.tex
\section{Conclusion}
\label{sec:conclu}
In this paper, we present \emph{DeepInv}, a novel self-supervised framework for diffusion inversion that for the first time enables accurate and efficient inversion through a trained end-to-end solver. 
Extensive experiments demonstrate that DeepInv outperforms existing inversion methods in both reconstruction quality and computational efficiency. Its integration with existing end-to-end editing methods not only improves output quality but also offers a promising direction for highly controllable diffusion-based real image editing.